\pdfoutput=1

\documentclass[11pt]{article}

\usepackage[final]{acl}

\usepackage{times}
\usepackage{latexsym}

\usepackage[T1]{fontenc}

\usepackage[utf8]{inputenc}

\usepackage{microtype}

\usepackage{inconsolata}

\usepackage{graphicx}
\usepackage{enumitem}
\usepackage{algorithm}
\usepackage{algpseudocode}
\usepackage{amsmath}
\usepackage{pgfplots}
\DeclareMathOperator*{\argmax}{arg\,max}

\pgfplotsset{compat=1.18}

\title{iPrOp: Interactive Prompt Optimization for\\ Large Language Models with a Human in the Loop}

\author{Jiahui Li \and Roman Klinger \\
  Fundamentals of Natural Language Processing, University of
  Bamberg, Germany \\
  \texttt{\{jiahui.li,roman.klinger\}@uni-bamberg.de}
}

\begin{document}
\maketitle
\begin{abstract}
  Prompt engineering has made significant contributions to the era of
  large language models, yet its effectiveness depends on the skills
  of a prompt author. This paper introduces \textit{iPrOp}, a novel
  interactive prompt optimization approach, to
  bridge manual prompt engineering and automatic prompt
  optimization while offering users the flexibility to assess evolving
  prompts. We aim to provide
  users with task-specific guidance to enhance human engagement in the
  optimization process, which is structured through prompt variations,
  informative instances, predictions generated by large language
  models along with their corresponding explanations, and relevant
  performance metrics. This approach empowers users to choose and
  further refine the prompts based on their individual preferences and
  needs. It can not only assist non-technical domain experts
  in generating optimal prompts tailored to their specific tasks or
  domains, but also enable to study the intrinsic parameters that
  influence the performance of prompt optimization. The evaluation
  shows that our approach has the capability to generate improved
  prompts, leading to enhanced task performance.
\end{abstract}
\begin{figure*}[ht]
  \centering
  \includegraphics[width=0.85\linewidth]{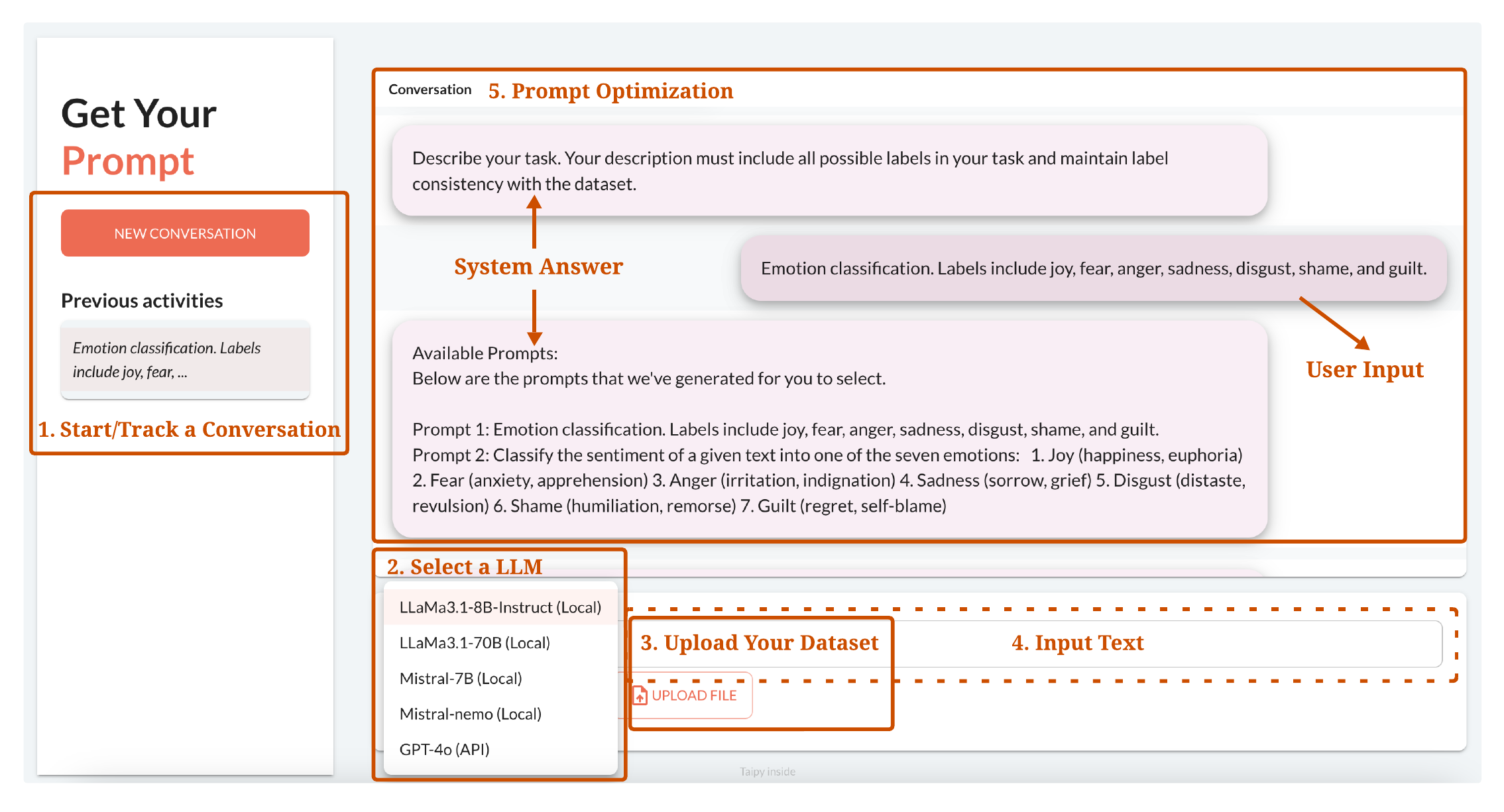}
  \caption{Screenshot of the \textit{iPrOp} Web application, where key components are annotated.}
  \label{fig:screen}
\end{figure*}

\section{Introduction}

With the advancement of large language models (LLMs), prompt engineering emerged for instructing these models to generate responses that align with users' requirements. Prompting allows LLMs to perform user-specified tasks, including tasks in previously unseen scenarios or particular domains \citep{devlin-etal-2019-bert,DBLP:journals/jmlr/RaffelSRLNMZLL20,mishra-etal-2022-cross}. 

However, prompt-based natural language processing (NLP) has
demonstrated limited robustness across domains, instances, or label
schemes
\citep{plaza-del-arco-etal-2022-natural,yin-etal-2019-benchmarking,zhou-etal-2022-prompt}. It
is also challenging to develop reliable methods for evaluation of LLMs
that factor in prompt brittleness \citep{ceron-etal-2024-beyond}. The
question of how to design a well-crafted prompt has received an
increasing amount of attention. Although there exists research on
analyzing which prompts are more effective for tasks like
classification and question answering
\citep{liu-etal-2022-makes,lu-etal-2022-fantastically,xu-etal-2022-exploring},
the need to efficiently identify high-quality prompts has sparked
increased attention into automatic prompt optimization
\citep{shin-etal-2020-autoprompt,pryzant-etal-2023-automatic}. However,
they tend to overlook the inherent contextuality and the domain-dependent nature of prompt engineering \citep{pei-etal-2025-selfprompt,claude3}. There is a lack of studies that combines user-guided prompt optimization with data-driven prompt optimization. Given that the user constitutes the ultimate authority to develop prompts that satisfy the varying trade-offs across different aspects of a specific task, we consider this an important research gap. 

Combining prompt optimization with a user in the loop comes with the
potential for a more guided engineering process, from which any user
may benefit. Two examples are particularly prominent:
(1) Technical laypeople
may require help with prompt development for dedicated
tasks. (2)~Manual prompt engineering may lead to biased
configurations, as generic prompts often fail to capture the complexities and nuances specific to particular domains, such as medical knowledge \citep{lu-etal-2023-medical}. Prior research has demonstrated the role of
human-in-the-loop methodologies in building robust systems
across a variety of tasks, including debugging text classifiers
\citep{lertvittayakumjorn-etal-2020-find}, hate speech classification
\citep{kotarcic-etal-2022-human}, and question answering chatbots
\citep{afzal-etal-2024-towards}.

To achieve the goal of supporting users in their prompt
development process, we hypothesize that a set of prompt properties is
important to decide if a prompt $p$ is considered better than another
prompt $p'$. These are (a) the performance of a prompt on some
annotated data, for instance measured by F$_1$ (we focus in this paper
on text classification tasks); (b) The readability and
interpretability of the prompt; (c) The quality of an explanation of
the predictions of the prompt; and (d), the alignment of the
annotations with the users expectations.  We therefore
propose an interactive prompt optimization approach with a
human-in-the-loop that considers all these aspects. The proposed
approach enables studies on the interaction between these various
parameters in the spirit of an iterative optimization in which the
automatic evaluation of an objective function is supported by a
human. We further envision that some decisions may be made
automatically, while others require the human to decide on the prompt
quality. Such collaborative decision process helps to maintain the
high quality of the prompts, while limiting the required user
interactions to those of particularly high value.

The repository of a prototypical web interface for the \textit{iPrOp} approach and an explanation video is
available at
\url{https://www.uni-bamberg.de/nlproc/ressourcen/iprop/}. Figure~\ref{fig:screen} presents a screenshot of the web-based user interface.

\begin{figure*}[t]
  \centering
  \includegraphics[width=0.8\linewidth]{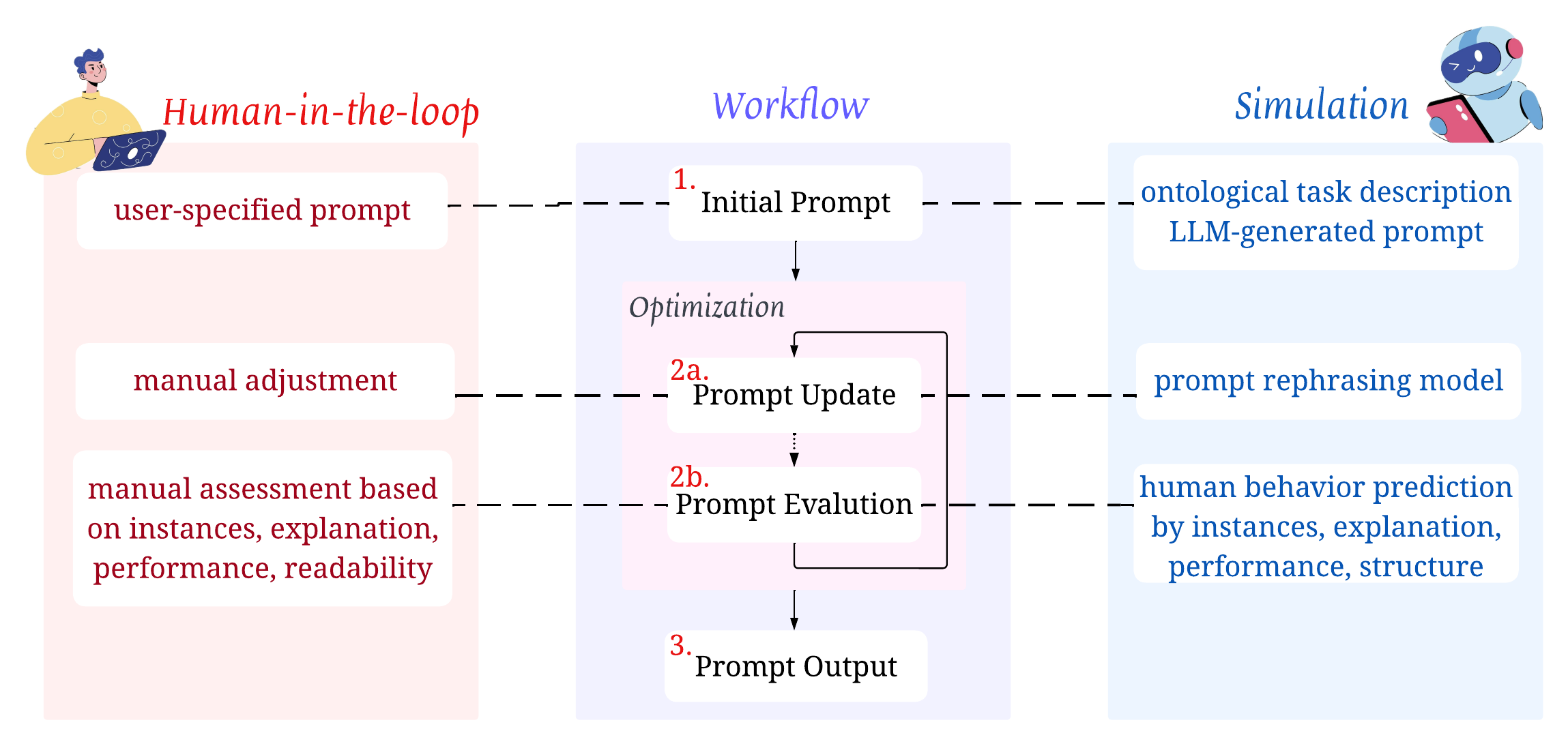}
  \caption{The conceptual workflow of our \textit{iPrOp} approach. The general workflow is shown in the middle. The left part shows potential human interaction in the various modules. To limit the amount of user interactions, each module can be supported by a simulated interaction.}
  \label{fig:concept}
\end{figure*}

\section{Related Work}

\subsection{Prompt Engineering for LLMs}
Prompt engineering is the process of designing and optimizing prompts to guide a language model for effective results on a downstream task. \citeposs{DBLP:journals/csur/LiuYFJHN23} survey categorizes previous works in prompt shapes and human-designed prompt templates. While the former category includes techniques such as cloze prompts \citep{cui-etal-2021-template} and prefix prompts \citep{li-liang-2021-prefix}, the latter focuses on manually crafted prompts \citep{DBLP:journals/corr/abs-2005-14165} and automated prompt templating processes \citep{shin-etal-2020-autoprompt}. Our work is derived from the latter case with the addition of human interventions.

The output of an LLM is influenced by the quality of prompts
\citep{lu-etal-2022-fantastically}. Prompts need to be adapted to
particular domains
\citep{karmaker-santu-feng-2023-teler,DBLP:conf/nips/WeiXM21}, and for
different LLMs \citep{chen-etal-2023-mapo}. Previous work therefore
attempted to search through paraphrases of prompts
\citep{jiang-etal-2020-know}, by compiling prompts based on templates
and class-triggering tokens \citep{shin-etal-2020-autoprompt}, or by
learning soft prompts \citep{qin-eisner-2021-learning}. Another
approach is to combine gradient descent method with hard prompts
\citep{DBLP:conf/nips/WenJKGGG23,pryzant-etal-2023-automatic}. In
contrast, our framework focuses on multiple factors such as task
selection, choice of LLM, and user-provided feedback as external
parameters. Further, we exploit the capabilities of LLMs as prompt
engineers
\citep{DBLP:conf/iclr/ZhouMHPPCB23,ye-etal-2024-prompt,DBLP:conf/icml/FernandoBMOR24,MenchacaResendiz2025}.

\subsection{Cooperative Artificial Intelligence}
This work is related to the field of cooperative artificial
intelligence, which touches upon topics of human-machine interaction
and efficient protocols of information exchange, enabling humans to
solve tasks collaboratively with machines. Such methods also
influenced NLP tasks, such as question answering
\citep{benamara-saint-dizier-2003-webcoop}, information retrieval
\citep{DBLP:books/daglib/0021593}, and chatbot interactions
\citep{hancock-etal-2019-learning}. More recent papers draw their
attention on collaborative annotation processes and model direct
manipulation
\citep{DBLP:journals/ki/BaurHLWVSA20,wang-etal-2021-putting}. However, we introduce a human-in-the-loop via replacing the automatic evaluation of an objective function by a human. Prior research has explored incorporated human feedback by presenting users with responses generated from paired prompts and asking for their preferences \citep{DBLP:journals/corr/abs-2405-17346}. In contrast, our framework offers a more comprehensive structure, encompassing a broader range of factors that should be considered during human evaluation.

\subsection{Explainable Artificial Intelligence}
Users which manually change properties of a system benefit from a good
understanding of the model's decisions. This task is approached by
explainable artificial intelligence (XAI) techniques
\citep{DBLP:journals/access/RoscherBDG20}. One prominent work that
introduced the interaction between model intervention and XAI is
\citet{teso_explanatory_2019}. Another study combines explanatory
interactive machine-learning methods with fair machine learning for
the bias-mitigation problem
\citep{DBLP:journals/make/HeidrichSSS23}. They both integrate
interpretability methods for machine learning models, such as SHAP
\citep{DBLP:conf/nips/LundbergL17}, LIME
\citep{DBLP:conf/kdd/Ribeiro0G16}, and Anchors
\citep{DBLP:conf/aaai/Ribeiro0G18}.

Although these tools offer intuitive explanations for classifiers,
their reliance on perturbations makes them computationally expensive
to apply to LLMs because of the high-dimensional nature and complexity
of LLMs. An alternative is to leverage the inherent explainability of
LLMs \citep{DBLP:journals/corr/abs-2401-13110}.
\citeposs{DBLP:journals/corr/abs-2403-08946} analysis of strategies to
enhance the transparency of LLMs. \citet{Bills2023} demonstrate that
LLMs are able to explain individual neurons in LLMs. This work
motivates our attempt to prompt LLMs for the explanations of their
predictions.

\section{Methods} \label{sec:method}

Figure~\ref{fig:concept} visualizes the conceptual workflow of our
\textit{iPrOp} approach. The workflow begins with an initial seed prompt
and proceeds through iterations of prompt updates and evaluations, led
by informative samples, explanations, and data evaluation with
performance metrics. To reduce human workload, each step can, in
principle, be performed either by the user or automatically. 

We formalize the process of the workflow as follows. The user is presented prompts in iterations and selects the preferred prompt $p^*$ based on their assessment $H$:
\[
    p^* = \argmax_{p \in P\cup M(P)}H(I(p_i)),
\]
Here, $M(P)$ is a prompt paraphrasing model that varies the prompts
$P$ selected from the previous iteration. $I(p_i)$ is a presentation
of prompt properties to the user, which consists of
\[
I(p_i) = (p_i, T^{p_i}_\alpha, E(T_\alpha,p_i), F_1(T^{p_i}_\beta)))\,.
\]

The user provides a (potentially small) training set $T$ for their
task, from which we sample two subsets $T_\alpha \subseteq T$ and
$T_\beta \subseteq T$ according to strategies $\alpha,
\beta$. $T^{p_i}_\alpha$ consists of instances to be shown to the user
together with model based explanations $E(T_\alpha,p_i)$. $T_\beta$
serves to calculate an evaluation score
$F_1(T^{p_i}_\beta)$ (we focus on text classification tasks for
simplicity).

This procedure is also visualized in Figure~\ref{fig:concept}. The
initialization of seed prompts ((1) in Figure~\ref{fig:concept})
requires users to describe the task. In simulation scenarios, this
process can be substituted with an ontological task description or
prompts generated automatically by LLMs. Subsequently, the initial
prompts are passed to the optimization modules. In the prompt update
module (2a), prompts are paraphrased. As an example, this paraphrasing of
\textit{`Classification task with labels: joy and sadness.'} with a
meta-prompt of an LLM \textit{`Rephrase the following prompt'} may
lead to \textit{`Classify the emotion of text into joy and
  sadness.'}
  
In the prompt evaluation stage (2b), the
human in the loop assesses the prompt quality, as described
above. Figure~\ref{fig:table} further provides a prototypical display
of the relevant information for two prompts to be chosen from. In the current prototype interface, the explanations are automatically generated by prompting a LLM. For instance, the specific prompt used is: \textit{`In your answer, provide only the label you choose and the explanation of your choice.'}. Examples of the generated explanations during the evaluation process are provided in the Appendix~\ref{sec:appendix}. The
optimization process is terminated once the user is satisfied (3).

\begin{figure}[t]
  \centering
  \includegraphics[width=\columnwidth]{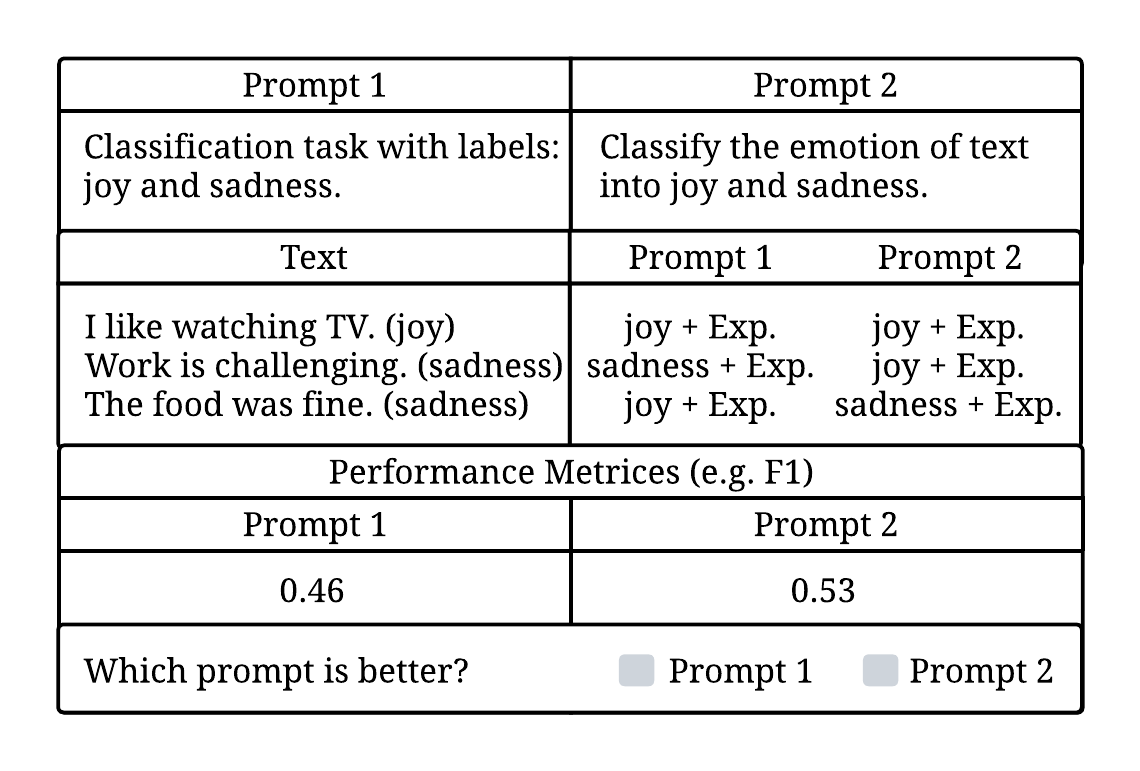}
  \caption{User interface prototype for an emotion analysis example during the interactive prompt optimization process. "Exp." refers to explanations for why a specific label is predicted by the model.}
  \label{fig:table}
\end{figure}
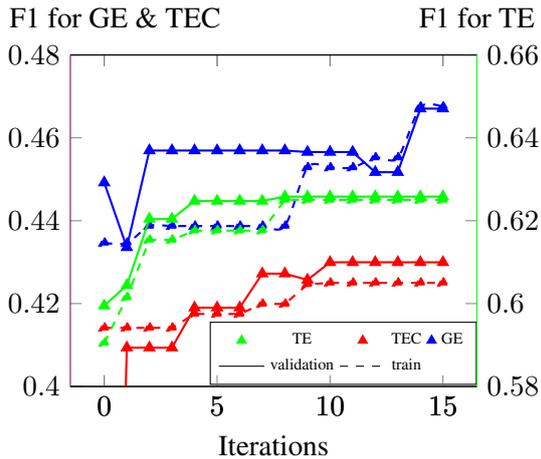
\begin{figure}[t]
    \centering
    
\begin{tikzpicture}
    \begin{axis}[
        xlabel={Iterations},
        ylabel={F1 for GE \& TEC},
        y label style={
                at={(0.11,1.05)}, 
                rotate=-90,
                anchor=south, 
            },
        width=0.9\columnwidth,
        legend style={at={(1,0.1)}, anchor=east, legend columns=3, font=\tiny},
        ymin=0.4, 
        ymax=0.48,
        axis y line*=left,
        y axis line style={violet},
    ]

    \addplot[only marks, mark=triangle*, color=green] coordinates {(0,0)};    \addlegendentry{TE}

    \addplot[only marks, mark=triangle*, color=red] coordinates {(0,0)};  
    \addlegendentry{TEC}

    \addplot[only marks, mark=triangle*, color=blue] coordinates {(0,0)};  
    \addlegendentry{GE}

    \addplot[only marks, smooth, color=black] coordinates {(0,0)};  
    \addlegendentry{validation} 

    \addplot[dashed, color=black] coordinates {(0,0)}; 
    \addlegendentry{train}  

    \addplot[
        smooth, thick, blue, dashed,mark=triangle*] 
        table[x=iter, y=getr, col sep=comma]{data.csv};

    \addplot[
        thick, blue,  
        mark=triangle*] 
        table [x=iter, y=geva, col sep=comma] {data.csv};
    \addplot[
        thick, red, dashed, 
        mark=triangle*] 
        table [x=iter, y=tectr, col sep=comma] {data.csv};

    \addplot[
        thick, red, 
        mark=triangle*] 
        table [x=iter, y=tecva, col sep=comma] {data.csv};
    \end{axis}

     \begin{axis}[
        ylabel={F1 for TE},
         y label style={
                at={(1,1.05)}, 
                rotate=-90,
                anchor=south, 
            },
        width=0.9\columnwidth,
        axis y line*=right,
        y axis line style={green},
        ymin=0.58, 
        ymax=0.66
    ]

    \addplot[
        thick, green, dashed,
        mark=triangle*] 
        table [x=iter, y=tetr, col sep=comma] {data.csv};
    \addplot[
        thick, green,  
        mark=triangle*] 
        table [x=iter, y=teva, col sep=comma] {data.csv};

    \end{axis}
\end{tikzpicture}
\caption{F1 scores for three datasets, shown separately on training and validation data. The abbreviations GE, TEC, and TE correspond to the \textsc{Grounded-Emotions} (blue), \textsc{TEC} (red), and \textsc{Tales-Emotion} (green) datasets, respectively. The left violet y-axis corresponds to \textsc{Grounded-Emotions} and \textsc{TEC}. The right green y-axis corresponds to \textsc{Tales-Emotion}.}
    \label{fig:f1}
  \end{figure}
\section{Evaluation}
\label{sec:eval}
We envision our \textit{iPrOp} approach to enable future research on the interaction of the various aspects to consider when humans make preference decisions on particular prompts under the available information. To validate the principled feasibility of our approach, we run
experiments on three emotion classification datasets using the
llama3.1:8b-instruct-fp16
model\footnote{\url{https://ollama.com/library/llama3.1:8b-instruct-fp16}}
\citep{llama3}. In this experiment, we only consider automated
classification performance scores and leave an automated evaluation of
the other measures or a user study for future work. In this
simulation, the prompt is selected corresponding to the weighted F$_1$
score over a fixed subset of the training data. We expect to
demonstrate a rising trend during the optimization process to verify
the effectiveness of our approach.

\paragraph{Datasets.} We select three datasets for single labeled
emotion classification task from \citet{bostan-klinger-2018-analysis},
namely \textsc{TEC}, covering general topics on tweets
\citep{mohammad-2012-emotional}; \textsc{Grounded-Emotions},
focusing on event-related topics on tweets \citep{DBLP:conf/acii/LiuBM17}; and \textsc{Tales-Emotion}, built upon fairytales \citep{tales_emotion}.

\paragraph{Result.} Figure~\ref{fig:f1} illustrates the F$_1$ scores
over 15 iterations. We observe an overall increasing trend in both
training and validation data.

\section{Conclusions and Future Work}
We proposed interactive prompt optimization as a novel approach to
configure instruction-tuned language models. The user is guided by information that is distilled from the
prompt and its performance on user-provided data. With this
approach, we suggested to aggregate information that may be relevant for
users to decide on prompt preferences.

The proposed approach has revealed several challenges that deserve
further investigation. There is a need to explore more effective
methodologies for enhancing the diversity of rephrased prompts. It is
important to limit the numbers of instances shown to the user, and that
selection requires methods to do so. It is essential to optimize the
various meta-prompts in the approach. Additionally, the optimization
algorithm is essential to improving the efficiency and
user-friendliness of our approach.

We envision that our \textit{iPrOp} approach lays the groundwork for
future research by addressing several open questions: (Q1) Which
parameters do influence the performance of the workflow configuration
in this approach? We presume that the example selection to better
understand how the prompt performs affects a user's ability to
estimate which prompt is preferable. Further, the methods to explain
the prompt prediction are crucial. Finally, underlying aspects such as
the model and its robustness are relevant factors for the approach to
succeed.  (Q2) How do prompts evolve throughout the optimization
iterations? An aspect of this question is what is the difference between automatic prompt
optimization and the human optimization is, and in which cases the
human intervention is indeed helpful. (Q3) To what extent can human
involvement be reduced while maintaining a balanced trade-off across
competing evaluation criteria? Can the interactive prompt optimization
approach be a collaborative learning procedure, in which the machine
only requests information if needed? We propose to study these
research questions based on the paradigm of interactive prompt
optimization introduced in this paper.

\section*{Limitations}

Although the \textit{iPrOp} approach offers a convenient interface
for non-technical users to attain suitable prompts, it has several
limitations that warrant consideration in the future
enhancement. First, in an effort to provide comprehensive explanations of LLM predictions, the challenge of computation time remains significant, and as a result, the streaming output is not effectively communicated to users. Second, developing an effective
strategy to address problems related to train-validation-test
splitting for user-provided datasets of varying sizes remains an
ongoing challenge. Third, the development of prompt optimization
iterations partially depends on the quality and variability of prompt
rephrasing. This implies that rephrased prompts may occasionally
retain low quality across multiple iterations. Furthermore, we observe
that certain datasets exhibit limited sensitivity to divergent
prompts, allowing a simple or even naive initial prompt to achieve
superior performance.

\section*{Acknowledgments}

This paper is supported by the project INPROMPT (Interactive Prompt
Optimization with the Human in the Loop for Natural Language
Understanding Model Development and Intervention, funded by the German
Research Foundation, KL 2869/13-1, project number 521755488).

\section*{Ethical Considerations}

Our approach is designed with careful attention to ethical standards in
data usage, privacy, and compliance with the ACL Code of Ethics. Our
method does not contribute to the republication or redistribution of
any datasets. The datasets used for testing and evaluation are
publicly available and we ensure that they have been collected
according to ethical standards before using them. To safeguard user
privacy, all data provided by users is stored exclusively on their
local machines. While potential risks associated with the underlying
LLMs could result in the exposure of user-provided datasets, we aim to
mitigate these risks by offering more secure local models. In
addition, our approach cannot guarantee that the optimal prompts
identified are state of the art for specific tasks. Furthermore,
individual preferences may introduce biases, which could potentially
mislead users. We are committed to continuously monitoring and
improving the ethical performance of our approach.

\bibliography{custom}

\newpage
\onecolumn
\appendix
\section{Appendix}
\label{sec:appendix}

\begin{table}[ht]
\centering
\small 
\begin{tabular}{p{6cm}p{8cm}} 
\hline
\textbf{Text} & \textbf{Predicted Label} \& \textbf{Explanation} \\
\hline
hey you! Hope school is going well for you! Love ya$\heartsuit$ & Joy.  The presence of the $\heartsuit$ symbol and the affectionate message ("Love ya") convey a positive emotional tone, indicating joy. \\ 
Mmmm breakfast in bed :) & Joy.  The person is expressing excitement and delight through the use of a happy emoticon :). \\
A wonderful story about how an act of kindness can change someone's life. & Joy.  The use of "wonderful" to describe the story suggests a positive tone, and the sharing of it on social media implies that the act of kindness is uplifting and inspiring. \\
Someday.... & Sadness.  The use of the word "someday" can imply a sense of longing or wistfulness, and the fact that it\'s used in conjunction with two links to online content may suggest that the author is reflecting on something they\'ve written in the past. \\
With immigration drop, who needs a wall? & Sadness.  The text mentions an "immigration drop", which implies a negative change or decrease in something related to immigration. This is likely to evoke feelings of sadness or concern rather than joy. The use of the phrase "who needs a wall" also suggests a sense of pessimism or hopelessness, further indicating that the overall tone of the text is more negative and therefore classified as sadness. \\
do not try to stop them & Sadness.  The use of "do not try to stop them" is a warning or cautionary statement, often implying something bad or undesirable is happening. The link provided does not provide context, but the tone of the message suggests a sense of urgency and possibly desperation, which are common indicators of sadness or concern. \\
\hline
\end{tabular}
\caption{Examples of an explanation generated for a predicted label in the \textsc{Grounded-Emotions} dataset task.}
\label{tab:explanation_examples}
\end{table}

\end{document}